\documentclass[journal]{IEEEtran}

\usepackage{cite}
\usepackage{graphicx}
\usepackage{multicol}
\usepackage{subfig}
\usepackage{amsmath}
\usepackage{amsfonts}
\usepackage{romannum}
\usepackage{caption}
\usepackage{tabularx}
\usepackage{tabulary}
\usepackage{array}
\usepackage{tabu}
\usepackage{textcomp}
\usepackage{xcolor}

\usepackage[ruled, vlined]{algorithm2e}
\SetKwComment{Comment}{$\triangleright$\ }{}
\DeclareMathOperator*{\argmax}{arg\,max}  
\DeclareMathOperator*{\argmin}{arg\,min}  

\ifCLASSINFOpdf
\else
\fi

\begin{document}

\title{Automatic Registration between Cone-Beam CT and Scanned Surface via Deep-Pose Regression Neural Networks and Clustered Similarities}

\author{Minyoung Chung, Jingyu Lee, Wisoo Song, Youngchan Song, Il-Hyung Yang, Jeongjin Lee$^{\ast}$, and Yeong-Gil Shin%
\thanks{\textit{Asterisk indicates corresponding author.}}%
\thanks{M. Chung, J. Lee, Y. Song, and Y.-G. Shin are with the Department of Computer Science and Engineering, Seoul National University, Korea (e-mail: chungmy@cglab.snu.ac.kr).}%
\thanks{W. Song is with the Institute for Applied Computational Science, Harvard University, MA 02138 USA (e-mail: victor\_song@g.harvard.edu).}%
\thanks{I.-H. Yang is with the Department of Orthodontics, Seoul National University School of Dentistry (e-mail: drortho@snu.ac.kr).}%
\thanks{*J. Lee is with the Department of Computer Science and Engineering, Soong-sil University, Korea (e-mail: leejeongjin@ssu.ac.kr).}}


\maketitle

\begin{abstract}
Computerized registration between maxillofacial cone-beam computed tomography (CT) images and a scanned dental model is an essential prerequisite in surgical planning for dental implants or orthognathic surgery. We propose a novel method that performs fully automatic registration between a cone-beam CT image and an optically scanned model. To build a robust and automatic initial registration method, our method applies deep-pose regression neural networks in a reduced domain (i.e., 2-dimensional image). Subsequently, fine registration is performed via optimal clusters. Majority voting system achieves globally optimal transformations while each cluster attempts to optimize local transformation parameters. The coherency of clusters determines their candidacy for the optimal cluster set. The outlying regions in the iso-surface are effectively removed based on the consensus among the optimal clusters. The accuracy of registration was evaluated by the Euclidean distance of 10 landmarks on a scanned model which were annotated by the experts in the field. The experiments show that the proposed method's registration accuracy, measured in landmark distance, outperforms other existing methods by 30.77\% to 70\%. In addition to achieving high accuracy, our proposed method requires neither human-interactions nor priors (e.g., iso-surface extraction). The main significance of our study is twofold: 1) the employment of light-weighted neural networks which indicates the applicability of neural network in extracting pose cues that can be easily obtained and 2) the introduction of an optimal cluster-based registration method that can avoid metal artifacts during the matching procedures.
\end{abstract}


\IEEEpeerreviewmaketitle

\section{Introduction}
\IEEEPARstart{C}{omputerized} registration of medical images is a challenging problem for many clinical applications (e.g., follow-up studies, surgical planning, and augmented reality \cite{alam2017medical, haas2015computer, cevidanes2005superimposition, stokbro2016surgical}). In dental application, registration between maxillofacial cone-beam computed tomography (CT) images and a scanned model is an essential prerequisite in surgical planning for dental implants or orthognathic surgery \cite{jamjoom2018positional, flugge2017registration, eggers2006image, plooij2011digital, gateno2007clinical, centenero20123d, ritto2018comparison, rangel2018integration, flugge2017registration, cevidanes2005superimposition, popat2010new}. Rigid transformation, which has a relatively small number of parameters, can be a viable method in the application. However, the registration problem between two different imaging protocols (i.e., optically scanned surface model and CT image) is a challenging task \cite{flugge2017registration}. Moreover, the metal artifacts in cone-beam CT images hinder the accuracy of the registration \cite{becker2018impact}. Manual registration or initial registration between CT and scanned dental model is a time-consuming task. Most of the clinical applications require the three points for the initial registration. The term ``three-points" refers to the registration using manual three points depiction. Thus, even the state-of-the-art clinical applications relying on the three-points step cannot be fully automated.\par

Many methods have been proposed to address the rigid registration problem in medical imaging applications. Intensity-based methods attempt to attain optimal registration parameters (i.e., transformation matrix) by intensity similarity criteria \cite{zitova2003image}. On the other hand, feature-based methods attempt to extract useful features (e.g., spatial points or descriptors) to match correspondences \cite{zitova2003image}. Iterative closest points (ICP) method \cite{besl1992method, rusinkiewicz2001efficient, bouaziz2013sparse} is widely used to perform point-to-point registrations. In dental field, the ICP method is an effective approach that can be applied to many applications such as real-time registration of optical scanning \cite{ahn2017development}. However, owing to the different protocols of the target and moving images, the ICP method is not directly applicable to the CT-model registration task. Some studies have been proposed to extract corresponding surface points from CT to apply ICP method with and without features \cite{herline2000surface, jin20063d, bolandzadeh2013multimodal, lin2013new, fan2014new, flugge2017registration, jung2018registration}. Markers (i.e., manual landmarks) have also been used to perform registration between CT and optically scanned dental model \cite{gateno2003new, tsuji2006new, yang2015automatic, rangel2012integration, fieten2009fast, de2016novel, becker2018impact}. It is also clearly known that registration performance can be increased by confident region priors \cite{sun2018registration}. The major limitations of the previously proposed methods are twofold: 1) initial registration is either not automated or lacks robustness and 2) the performance of matching is seriously affected by the extent of metal artifacts. The performance cannot guarantee robustness in the majority of cases due to the presence of metal artifacts.\par

\begin{figure*}[t!]
    \centering
    \subfloat[Initial registration framework. The input mesh model and CT image are projected to 2D images. We apply principal component analysis to the mesh model to obtain the main axis of projection. Subsequently, a synthetic depth image is generated using the model by defining a tight bounding plane. In the case of the CT image, we project maximum intensities with respect to the x-axis. The two 2D images are then fed to the regression neural networks to acquire alignment cues, i.e., point (red dot) and line (green arrow). Finally, the mesh model and CT image are initially matched with point and line pairs.]{
    \includegraphics[width=\linewidth]{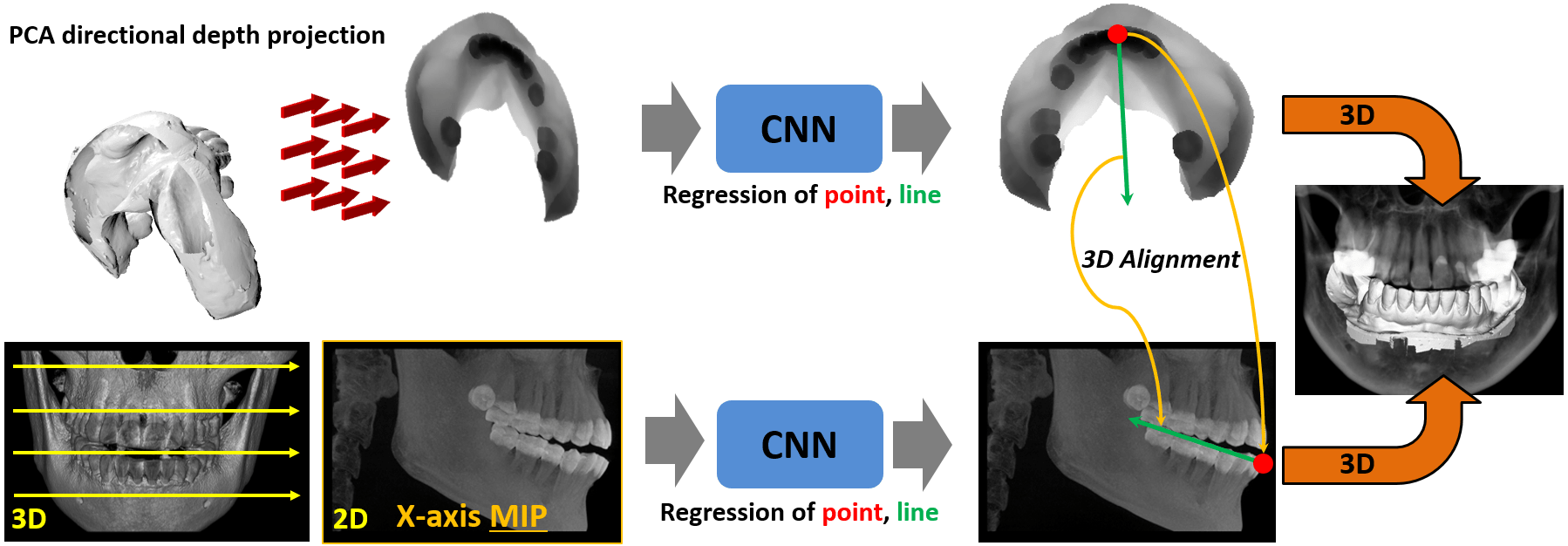}
    \label{fig:reg_initial}}
    
    \centering
    \subfloat[Fine registration using optimal clusters. From the initially matched state, local clusters are generated on a surface model for individual registration based on clusters. After local optimization, optimal clusters are selected based on the ``voting" results among cluster candidates. The final clusters are then used for final registration.]{
    \includegraphics[width=\linewidth]{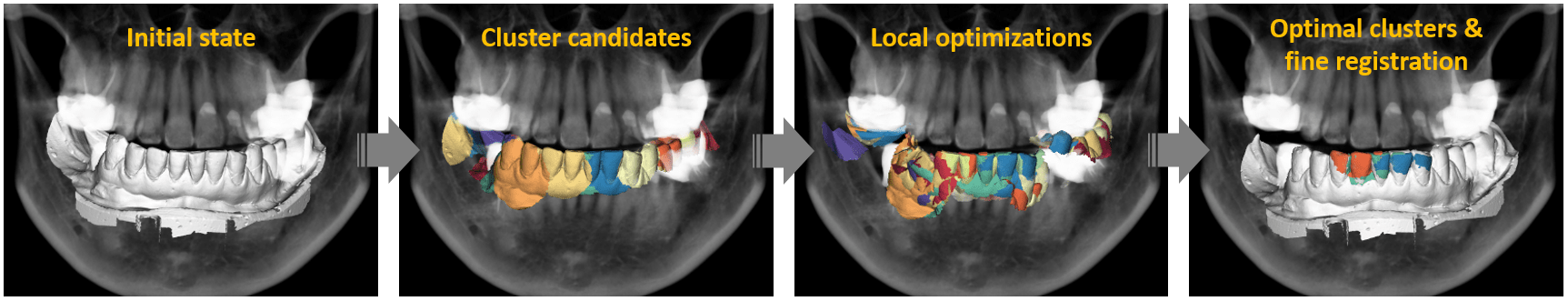}
    \label{fig:reg_final}}

    \caption{The overall architecture of the proposed algorithm. (a) The initial registration procedure using deep-pose regression neural networks. (b) Subsequent fine registration via clustered similarities. Best viewed in color.}
    \label{fig:reg_init}
\end{figure*}

In recent years, many studies have used deep neural network to resolve automatic registration problems in the field \cite{jiang2010medical, litjens2017survey, yang2016fast, simonovsky2016deep, miao2016cnn, yang2017quicksilver, hu2018weakly}. Some proposed unsupervised learning framework with certain similarity criteria (e.g., intensity and landmarks) and others proposed supervised \cite{yang2016fast, simonovsky2016deep} or weakly supervised neural network \cite{hu2018weakly} to perform registration. However, most of the previous works rely on intensity-based similarity or suffer from the deficiency of annotated data. The lack of ground-truth data makes it hard to train the neural network in a fully supervised manner. As aforementioned, although deep neural networks have been actively applied to various dental applications \cite{Hwang2019AnOO}, the manual depiction of three-points is still required for registration in the clinics.\par

In this paper, we propose a novel and robust method to perform fully automatic registration between cone-beam CT images and the scanned dental model without any fragile priors (e.g., iso-surface extractions and ICP). We first extract the alignment cues (i.e., pose) from the given CT and model via deep convolutional neural networks (CNNs) followed by the rough model alignment. Finally, fine matching is performed by optimal clusters that are obtained according to the similarity measurement. The key achievements of our method are: fully automated initial alignment and robust fine registration results independent of the presence of metal artifacts in the CT images.\par

The remainder of this paper is structured as follows. In Section {\Romannum{2}}, we will describe our proposed method in detail. Section {\Romannum{3}}, {\Romannum{4}}, and {\Romannum{5}} illustrate the experimental results, discussion, and conclusion, respectively.


\section{Methodology}
Our method consists of two steps: 1) deep-pose regression and 2) optimal cluster-based matching (Fig. \ref{fig:reg_init}). The first step aligns the model to CT in rough conditions (i.e., initial registration). The second matching procedure performs fine registration via optimal clusters. The details of each method are described in the following sections.

\begin{figure*}[t]
    \centering
    
    \subfloat[Schematic view of cluster transformations and coherency. The local optimal transformation of cluster $C^i$ (i.e., $T_i$) is applied identially to other clusters for coherence error calculation in (\ref{eq:dist_error}). $C_0^*$ and $C_1^*$ are the first and second minimum coherency error-valued clusters.]{\includegraphics[width=3.1in]{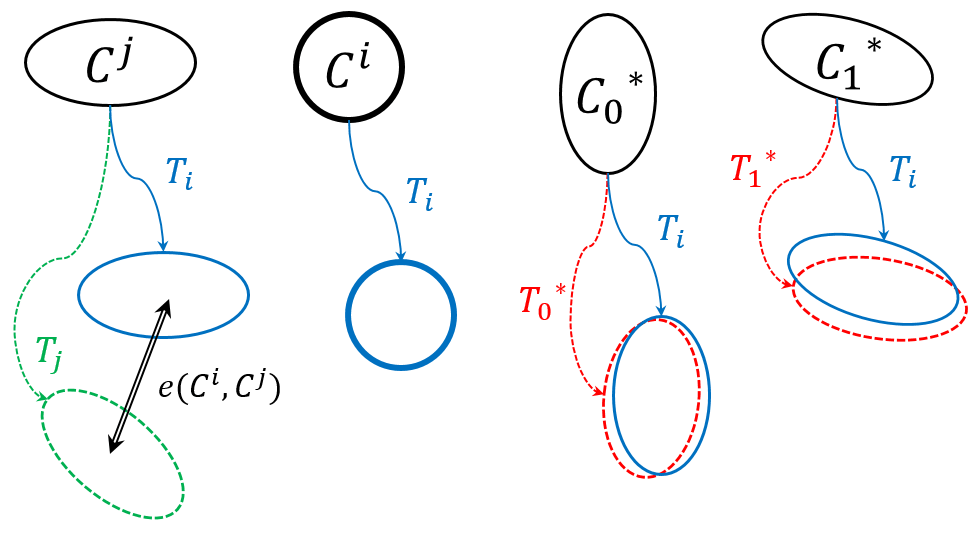}%
    \label{fig:coh0}}
    \hspace{0.15in}
    \subfloat[The red line (i.e., $v^{T_i}$) represents the maximum coherency vector (i.e., minimum coherency error vector derived from (\ref{eq:coherency})).]{\includegraphics[width=1.8in]{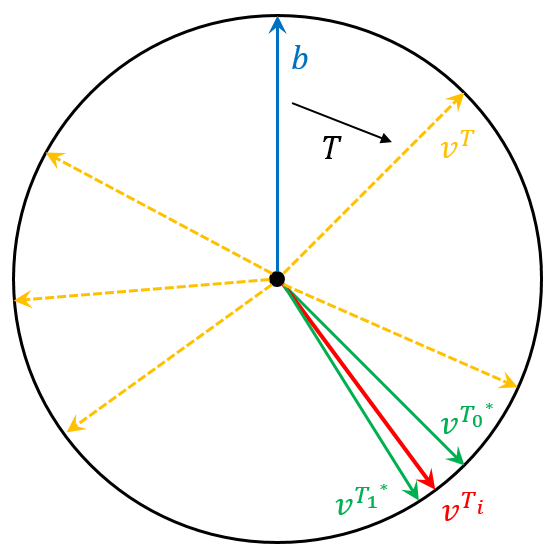}%
    \label{fig:coh1}}
    \hspace{0.15in}
    \subfloat[The red line (i.e., $v^{T_i}$) represents the minimum coherency vector (i.e., $C^{min}$) achieved by (\ref{eq:outlying}).]{\includegraphics[width=1.8in]{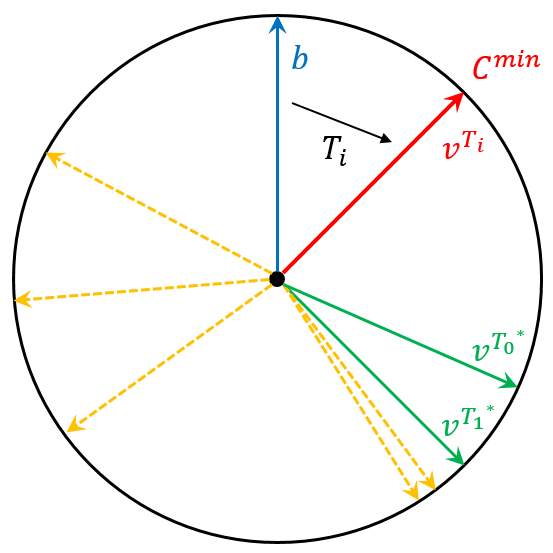}%
    \label{fig:coh2}}
    
    \caption{(a) shows the schematic view of cluster transformations. Each dotted oval represents transformed cluster obtained as a result of applying locally optimized transformations. (b) and (c) demonstrates the coherency evaluation with respect to 2D rotational transformation. Base vector $b$ represents every cluster for simplicity. Each red line (i.e., $v^{T_i}$) represents the given transformed vector. The green lines are the first and second transformed vectors that are most similar to each vector (i.e., $T_0^*$ and $T_1^*$ in (a)). The coherency error can be calculated as $c_i={||v^{T_i}-v^{T_0^*}||}_2+{||v^{T_i}-v^{T_1^*}||}_2$. Best viewed in color.}
    \label{fig:coherent}
\end{figure*}

\subsection{Deep-Pose Regression}
In this step, initial registration between CT and the model is performed. Directly using 3-dimensional (3D) information is too complex and noise-dependent. Therefore, we reduced the dimension to 2D for a robust initial alignment. Let CT image $I(\textbf{x})\longrightarrow\mathbb{R}$, where $\textbf{x}\in\Omega$ ($\Omega\in\mathbb{R}^3$). Maximum intensity projection (MIP) image with respect to x-axis direction, $I_{p}$, is generated from $I$ (indicated by yellow arrows in Fig. \ref{fig:reg_initial}). As for the scanned model, a synthetic depth image is generated for the main axis (i.e., full-arch visible axis). The main axis can be easily obtained by principal component analysis (PCA) \cite{horn1990matrix, wold1987principal}. In a given triangular mesh model $M=\{V,E\}$, where $V$ and $E$ are sets of vertices and edges, let a 3D vector $\textbf{v}\in{V}$ be a positional vector in set $V$. Defining the mean vector by $u=\frac{1}{|V|}\sum_i v_i$, the covariance matrix of an input data can be defined by $C=\frac{1}{|V|}\sum_i (v_i-u)(v_i^T-u^T)$. PCA analysis can be subsequently performed via eigen-decomposition or singular value decomposition \cite{horn1990matrix, golub1970singular, strang1993introduction}:
\begin{equation}
    Cv_i=\lambda_iv_i, i\in \{0, 1, 2\},
\label{eq:pca}
\end{equation}
where $\lambda_i$ are the eigenvalues of $C$ ($\lambda_0>\lambda_1>\lambda_2$). Depth image, $M_{p}$, is then generated by projecting all vertices to a tight bounding plane that has $v_2$ as a normal vector. Images are normalized to the range of [0..1]. Then we use the trained CNN models to acquire corresponding point and line pairs for each image (Fig. \ref{fig:reg_initial}). The point and line pairs are subsequently reconstructed (i.e., positioned) in the 3D domain. Since the projected bounding plane, with respect to the scanned model, is originally defined in a 3D domain, the points and lines in $M_p$ are automatically positioned in the 3-dimensional space. In case of $I_p$, the x-coordinate is set to $w/2$ (i.e., the center point for x-axis) where $w$ is the width of an input CT image. Finally, the model is spatially matched to the 3D CT image with 3D points and lines (i.e., overlapping the bounding plane of the model and a corresponding plane in a CT which is sliced by a vector in $I_p$ (Fig. \ref{fig:reg_initial})). Whether the scanned model is the upper jaw (i.e., maxilla) or the lower jaw (i.e., mandible) is given as a prior.\par
For training, we used $200$ scanned models and $100$ CT images. We manually annotated (i.e., point and angle of the line) images for $M_p$ and $I_p$. The overall loss is formulated as follows.
\begin{equation}
    \chi(X, \mathbf{p}, \theta; W)={||\mathbf{p}-\mathbf{y_1}||}_2+\alpha{||\theta-y_2||}_2+\beta{||W||}_2^2
\label{eq:cnn}
\end{equation}
where $X$, $\textbf{p}$, and $\theta$ are the input 2D image, ground-truth 2D point and angle of the line, respectively. $W$ represents the weights of the network, $\mathbf{y_1}$ and $y_2$ are the network outputs. The network is trained according to the weighting parameters $\alpha$ and $\beta$. For training and inference, we used the traditional VGG-16 network \cite{simonyan2014very} with a slight modification in the final layer to output a 3D tensor (i.e., a 2D point and an angle). Two identical neural networks were used to train each projection image (i.e., $M_p$ and $I_p$). The only difference is the final output tensor which is 6D for $I_p$ (i.e., a pair of a 2D point and an angle). `Xavier' initialization \cite{glorot2010understanding} was used for initializing all the weights of the network. While training the network, we fixed the loss parameter as $\alpha=\beta=0.1$. We used the Adam optimizer \cite{kingma2014adam} with batch size 64 and set the learning rate to 0.001. We decayed the learning rate by multiplying 0.1 for every 20 epochs. We trained the network for 100 epochs using an Intel i7-7700K desktop system with 4.2 GHz processor, 32 GB of memory, and Nvidia Titan Xp GPU machine. It took 1h to complete all the training procedures.


\subsection{Optimal Cluster-based Matching}
\begin{figure}[t!b]
    \centering
    \subfloat[Regression on point (dot) and line (arrow) pairs for x-axis projected MIP images (i.e., $I_{p}$). There are two pairs for the lower (red) and the upper (blue) jaw.]{
    \includegraphics[width=\linewidth]{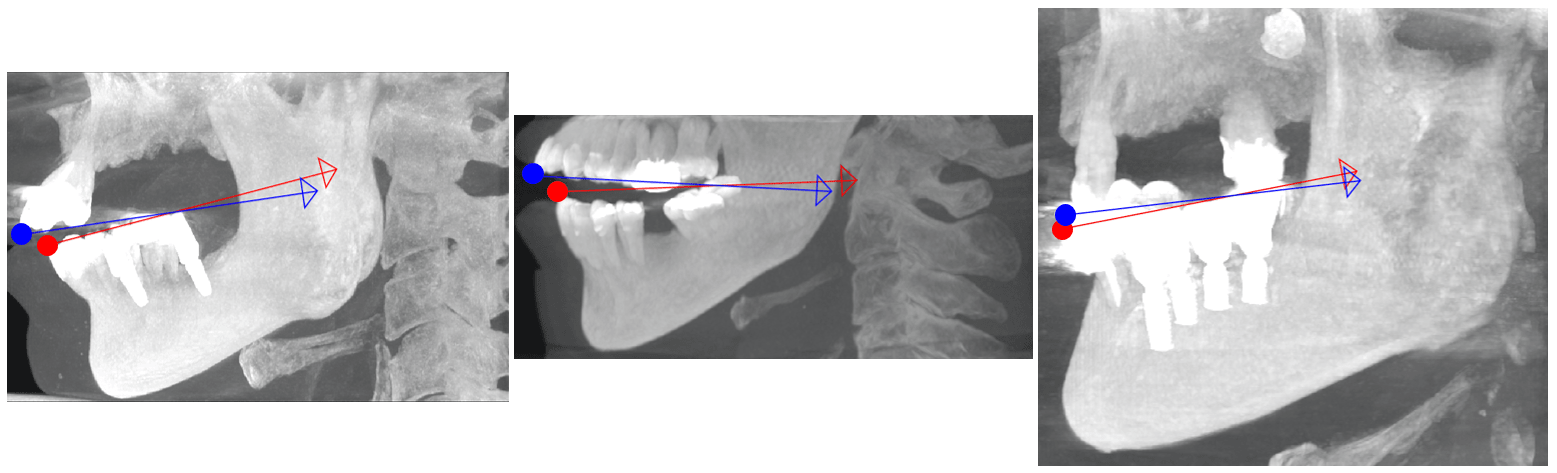}
    \label{fig:regress_mip}}
    
    \centering
    \subfloat[Regression on point and line for depth projected images of the lower jaw (i.e., $M_{p}$).]{
    \includegraphics[width=\linewidth]{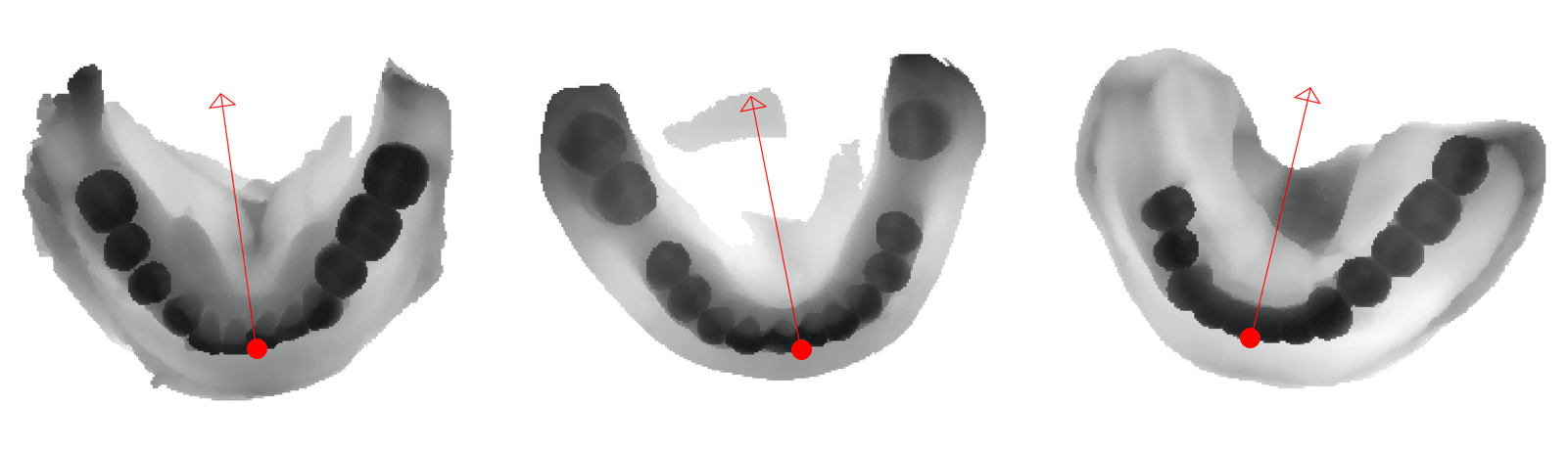}
    \label{fig:regress_depth}}
    
    \centering
    \subfloat[Regression on point and line for depth projected images of the upper jaw (i.e., $M_{p}$).]{
    \includegraphics[width=\linewidth]{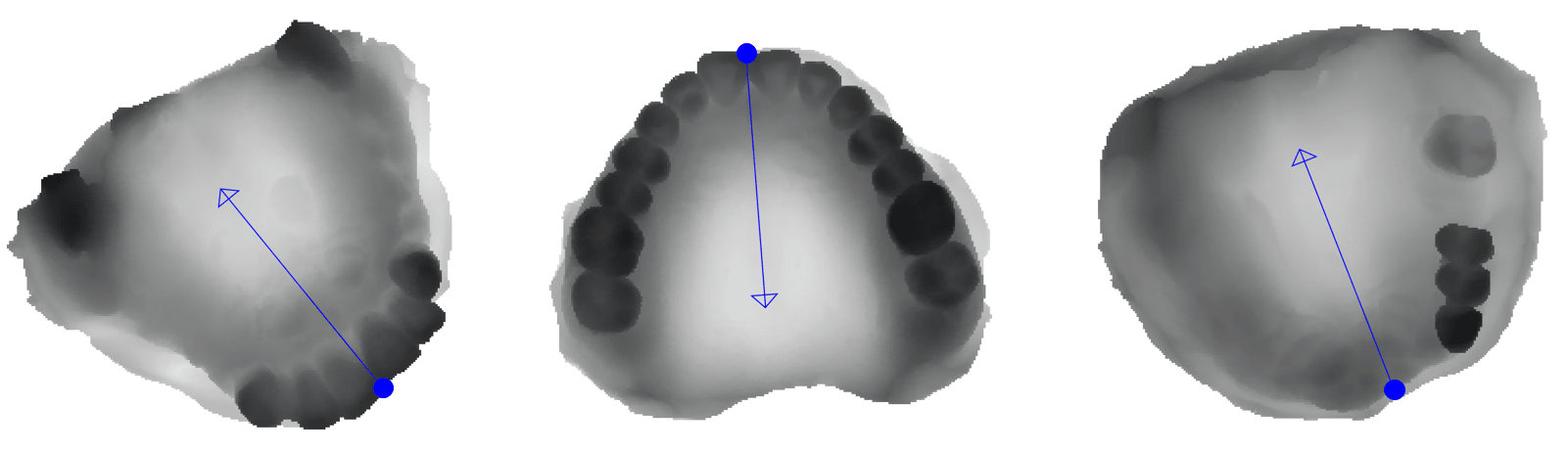}
    \label{fig:regress_depth2}}
    
    \centering
    \subfloat[Initial registration result. The two models (b, c) are matched to the CT image via point and line pairs.]{
    \includegraphics[width=\linewidth]{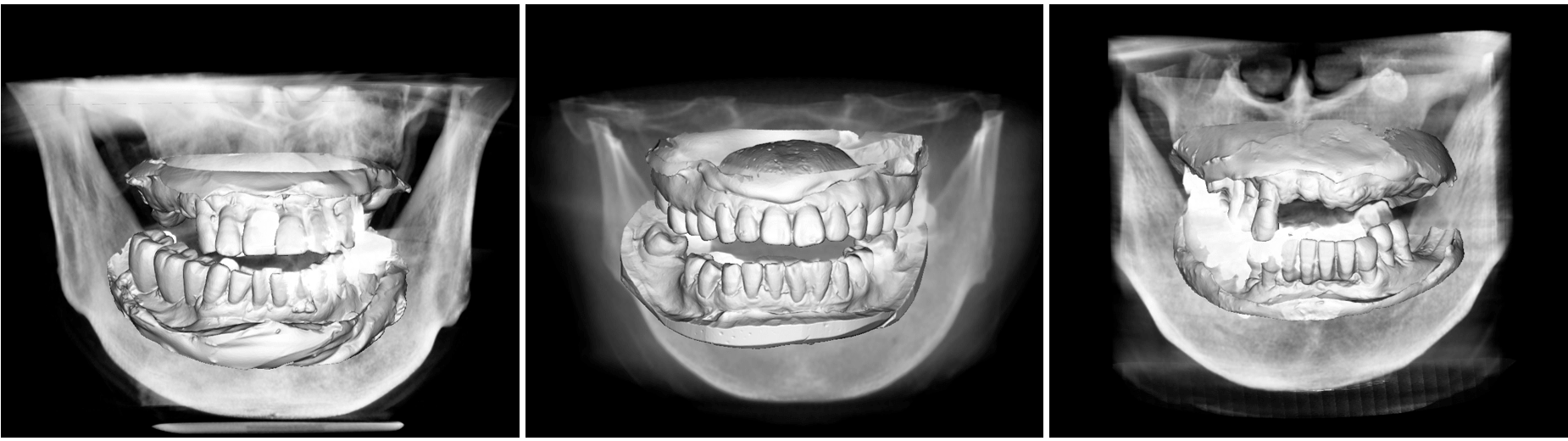}
    \label{fig:init_reg}}
\caption{Point and line regression results in (a) MIP and (b, c) depth projected images. The corresponding initial registration results are shown in (d).}
\label{fig:regress}
\end{figure}

Cluster $C_{v,r}\in{M}$ is defined by the center vertex $v$ and radius $r\in{\mathbb{R}}$, i.e., connected local vertices in $M$ constrained by $r$ centered at $v$. We use the notation $C$ in the remaining text for simplicity. Multiple clusters are automatically generated after initial registration (Fig. \ref{fig:reg_final}). We limited the minimum distance among the center of clusters to $r$ (i.e., allowing overlaps). Since we already know a projection vector of a given mesh model, we prioritized the vertices that are located in the crown region (i.e., lower values in $M_p$ image (Fig. \ref{fig:regress_depth})). Finally, we additionally added clusters in a stochastic manner. We added three clusters for each cluster by randomly positioning the center distanced less than $r$mm with $\pm10^{\circ}$ random rotations for each axis. The presented stochastic procedure enhances the stability of the results by means of improving the probability of the accurate local matching performance.\par

Local optimization (i.e., registration) is performed for each $i^{th}$ cluster ($i\in [0..N-1]$, where $N$ is the number of clusters) according to the vector alignment-based similarity criteria:
\begin{equation}
    L(M, I; C^i, T_{\theta_i})=-\sum_{\mathbf{v}\in C^i}<T_{\theta_i}^{-1T}(\overrightarrow{M(\textbf{v})}), \nabla I(T_{\theta_i}(\textbf{v}))>,
\label{eq:sm}
\end{equation}
where $\overrightarrow{M(\textbf{v})}$ is a normal vector at vertex $\textbf{v}$, $\nabla I(\textbf{v})$ is a gradient at $\textbf{v}\in\Omega$, and $T_{\theta_i}$ is a transformation with respect to the parameters, $\theta_i$. We used downhill simplex method \cite{lagarias1998convergence} for local cluster optimization until convergence within six dimensional parameters (i.e., three dimensional parameters for each translation and rotation; $\theta_i$):
\begin{equation}
    \argmin_{\theta_i}-L(M, I; C^i, T_{\theta_i}).
\label{eq:minimize}
\end{equation}
\par

\begin{algorithm}[t]
    \SetAlgoLined
    \caption{Selecting Three Optimal Clusters.}\label{alg:cluster}
        \KwIn{$I$, $M$, $C^i$, and $\theta_i$.}
        \KwOut{$C^*$}
        \KwData{Set of clusters, $C^*$}
        initialize $\theta_i \leftarrow I$, $n \leftarrow N$, $C^* \leftarrow \{C^i| 0 \leq i < N\}$\;
        \For{$i$, where $C^i \in C^*$}{
            $\theta_i \leftarrow \argmin_{\theta_i}-L(M, I; T_{\theta_i})$ \Comment*[r]{(\ref{eq:minimize})}
        }
        \BlankLine
        \While{$|C^*| > 3$}{
            $C^{min} \leftarrow$ \textit{Find\_Outlying\_Cluster($C^*$)} \Comment*[r]{(\ref{eq:outlying})}
            \BlankLine
            $C^* \leftarrow C^* - \{C^{min}\}$\;
        }
\label{algo}
\end{algorithm}

After performing local optimizations of the clusters, we have scattered transformation matrices for each cluster. Let $T_i$ represent the locally optimized global transformation matrix of a cluster, $C^i$. Defining the distance function of a cluster with respect to two different transformations as
\begin{equation}
    d(C; T_1, T_2)=\frac{1}{|C|}\sum_{\mathbf{v}\in C}{||T_1(\mathbf{v})-T_2(\mathbf{v})||}_2,
\end{equation}
the mutual coherency error for two clusters' transformation is approximated by the following equation:
\begin{equation}
    e(C^i, C^j)=d(C^j; T_i, T_j).
\label{eq:dist_error}
\end{equation}
The final coherency error for a given cluster is defined by
\begin{equation}
    c_i=e(C^i, C^*_0)+e(C^i, C^*_1),
\label{eq:coherency}
\end{equation}
where $C^*_0$ and $C^*_1$ represent the first and second minimum coherency error-valued clusters with respect to $C^i$ (Fig. \ref{fig:coh0}).
The outlying cluster is defined as
\begin{equation}
    C^{min}=\argmax_{C^i}c_i.
\label{eq:outlying}
\end{equation}
The schematic view of coherency evaluation is presented in Fig. \ref{fig:coherent}. Subsequently, we remove the worst coherently transformed cluster. The iteration continues until the optimal cluster set contains three clusters. The detailed procedures are illustrated in Algorithm \ref{algo}. The final three optimally chosen clusters, $C^*$, again iterate over the fine optimization process. In the fine registration stage, we integrate the three optimal clusters and iterate over the optimization process similar to (\ref{eq:minimize}):

\begin{equation}
    \theta^*=\argmin_{\theta}-L(M, I; C^*, T_{\theta}).
\label{eq:final}
\end{equation}

\section{Experiments}
In this section, we present a clear visualization of deep-pose regressions and initial registration results together with the optimal clusters. Accuracy and time complexity were the primary criteria for performance evaluation. Accuracy is evaluated with the ground-truth registration result obtained by the clinical experts in the field.\par
We acquired data from 145 subjects, each data sample including both a CT image and the scanned models of the subject's upper/lower jaw (i.e., 290 pairs for possible registration). The images are sourced from four different multi-centers. The model is acquired by optically scanning the surface from the cast of the patients. Each optically scanned surface formed a triangular mesh structure with $35-70$ thousand triangles and $18-40$ thousand vertices. In the CT dataset, the thickness of the slices ranged from 0.2 to 2.0mm and pixel sizes ranged from 0.2 to 1.0mm. We used 100 subjects for training, and 45 subjects for testing.\par

\begin{table}[tb]
\renewcommand{\arraystretch}{1.8}
\captionsetup{justification=centering, labelsep=newline}
\caption{MEAN AND STANDARD DEVIATION OF EUCLIDEAN DISTANCE ERROR FOR 8-FOLD CROSS-VALIDATION OF THE POINT AND LINE REGRESSIONS}
\label{table:results_stat}
\begin{tabularx}{\linewidth}{c||>{\centering\arraybackslash}X|>{\centering\arraybackslash}X}
\centering
Input & Point [mm] & Line [\textdegree]\\
\hline
$I_p$ & \(2.08\pm1.45\) & \(2.54\pm2.56\)\\
$M_p$ & \(8.01\pm6.20\) & \(2.59\pm2.37\)\\
\end{tabularx}
\end{table}

\begin{table}[tb]
\renewcommand{\arraystretch}{1.8}
\captionsetup{justification=centering, labelsep=newline}
\caption{THE MEAN EUCLIDEAN DISTANCE ERRORS OF LANDMARKS AND RUN-TIME}
\label{table:results_stat3}
\begin{tabularx}{\linewidth}{X||>{\centering\arraybackslash}X|>{\centering\arraybackslash}X}
\centering
Methods & Mean [mm] & Duration [s]\\
\hline
Three-points & \(2.73\pm0.51\) & N/A\\
Three-points\newline + ICP \cite{besl1992method} & \(6.29\pm2.71\) & $1.37\pm0.40$\\
Three-points\newline + Sparse ICP \cite{bouaziz2013sparse} & \(3.11\pm0.44\) & $23.11\pm4.14$\\
Three-points\newline + Go-ICP \cite{yang2016go} & \(6.30\pm2.68\) & $20.74\pm3.32$\\
Ours (only initial) & \(4.39\pm1.48\) & $3.67\pm0.06$\\
Ours & \(\textbf{1.89}\pmb{\pm}\textbf{0.82}\) & $17.68\pm7.76$\\
Ours\newline (w/o stochastic) & \(\textbf{2.37}\pmb{\pm}\textbf{0.56}\) & $7.14\pm1.99$\\
\end{tabularx}
\end{table}

\subsection{Deep-Pose Regressions}
In this section, we visualize the point and line regressions and the corresponding initial registration results. Fig. \ref{fig:regress} shows each result along with its corresponding $I_p$ and $M_p$. The initial registration results using points and lines are clearly shown in Fig. \ref{fig:init_reg}. For quantitative analysis, we used 8-fold cross validation metric for 100 subjects. Table \ref{table:results_stat} shows the Euclidean distance errors for each point and line. The result shows that there is no significant variation in the inference (i.e., testing). Even if the initial deep-pose regression resulted in a slight misalignment, the following optimal cluster-based fine registration step can complement the transformation biases.\par

\begin{figure}[t!]
    \centering
    \subfloat[Initial cluster candidates.]{\includegraphics[width=3.5in]{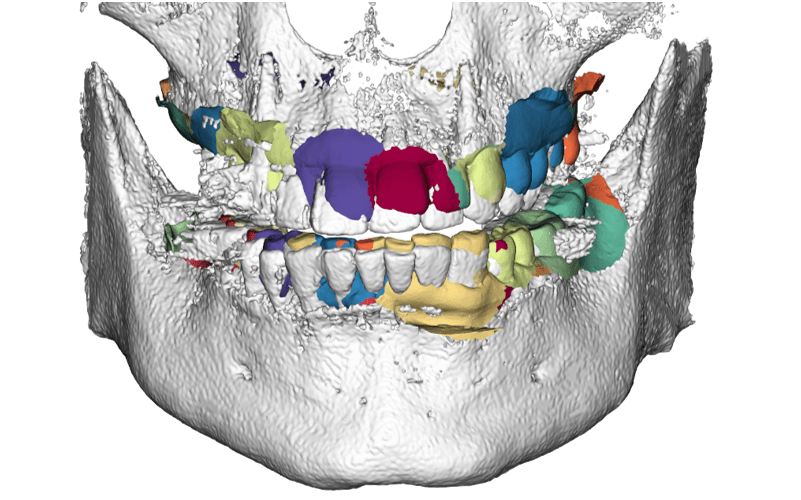}%
    \label{fig:cls1}}
    \vfil
    \subfloat[Locally optimized clusters.]{\includegraphics[width=3.5in]{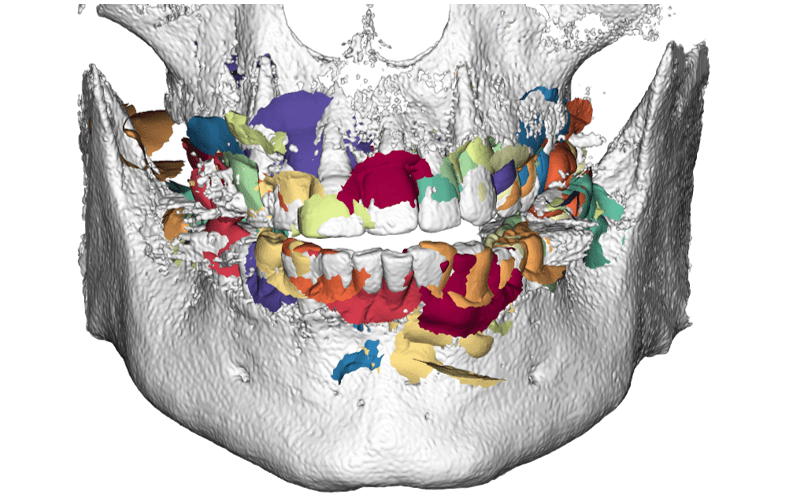}%
    \label{fig:cls2}}
    \vfil
    \subfloat[The final three optimal clusters.]{\includegraphics[width=3.5in]{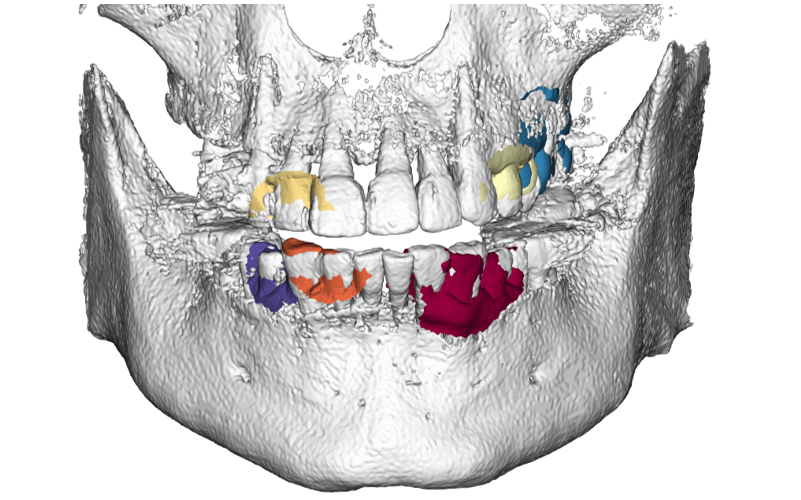}%
    \label{fig:cls3}}
    \caption{Visualization of clusters. (a) shows the initial clusters automatically generated by a given radius $r=10$mm. (b) shows the locally optimized clusters and (c) shows the final three clusters (i.e., the output of Algorithm \ref{algo}). The three clusters were selected by avoiding the metal artifacts. Best viewed in color.}
    \label{fig:cls}
\end{figure}

\begin{figure*}[tb]
    \centering
        \begin{minipage}[b]{1.33in}
            \captionsetup[subfigure]{labelformat=parens,labelsep=space,font=small}
            \centering
                \vfil
                \includegraphics[width=1.3in]{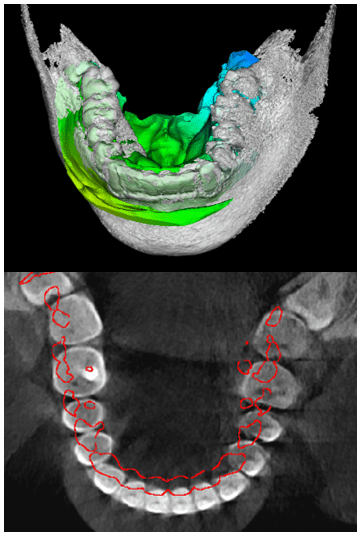}
                \includegraphics[width=1.3in]{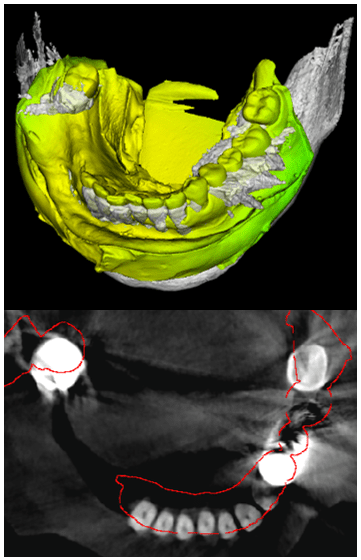}
                \includegraphics[width=1.3in]{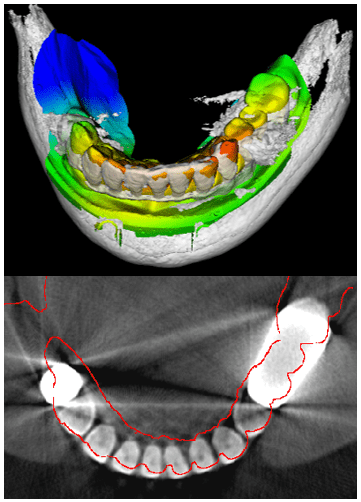}
            \captionof{subfigure}{Three-points.}
        \end{minipage}
        \hfil
        \begin{minipage}[b]{1.33in}
            \captionsetup{type=figure}
            \centering
                \vfil
                \includegraphics[width=1.3in]{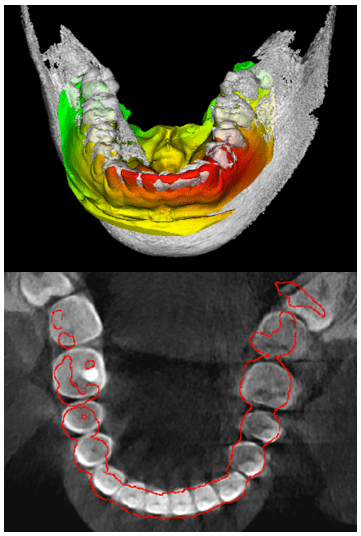}
                \includegraphics[width=1.3in]{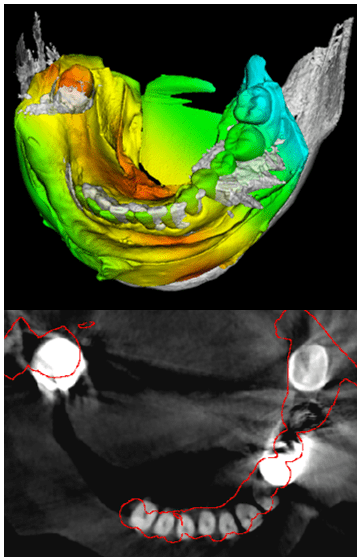}
                \includegraphics[width=1.3in]{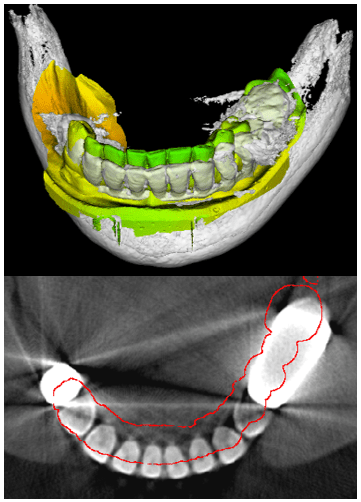}
            \captionof{subfigure}{+ICP.}
        \end{minipage}
        \hfil
        \begin{minipage}[b]{1.33in}
            \captionsetup{type=figure}
            \centering
                \vfil
                \includegraphics[width=1.3in]{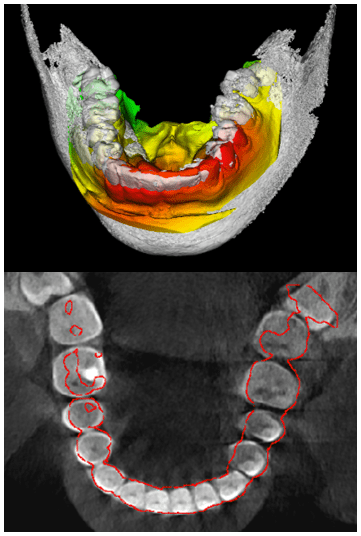}
                \includegraphics[width=1.3in]{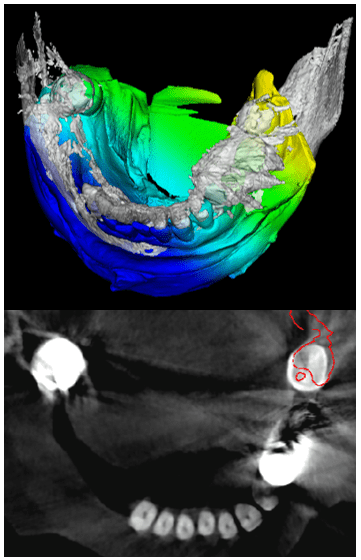}
                \includegraphics[width=1.3in]{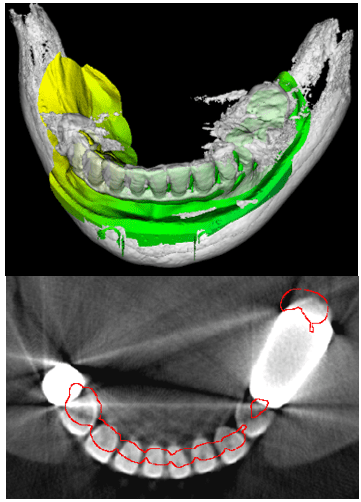}
            \captionof{subfigure}{+ Sparse ICP.}
        \end{minipage}
        \hfil
        \begin{minipage}[b]{1.33in}
            \captionsetup{type=figure}
            \centering
                \vfil
                \includegraphics[width=1.3in]{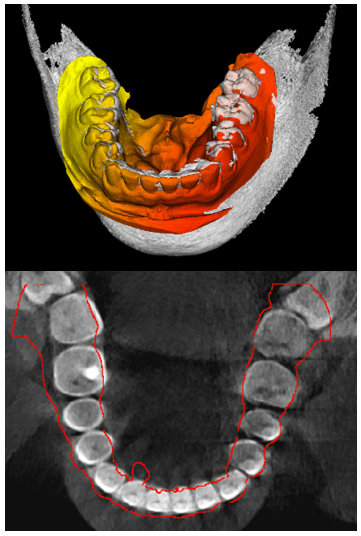}
                \includegraphics[width=1.3in]{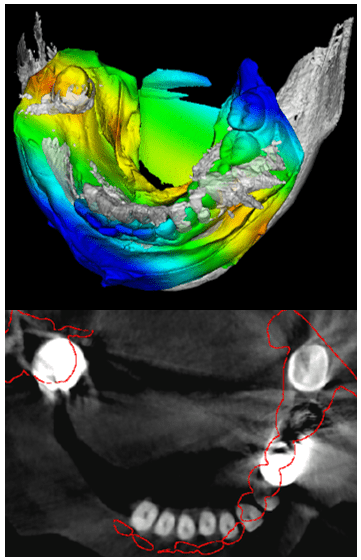}
                \includegraphics[width=1.3in]{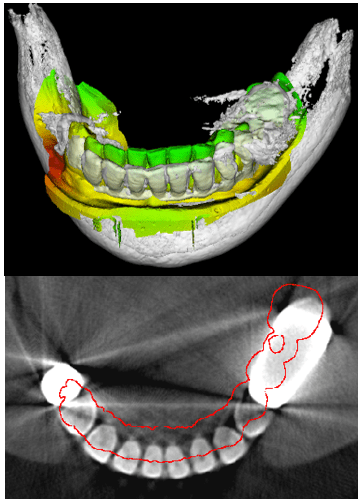}
            \captionof{subfigure}{+ Go-ICP.}
        \end{minipage}
        \hfil
        \begin{minipage}[b]{1.33in}
            \captionsetup{type=figure}
            \centering
                \vfil
                \includegraphics[width=1.3in]{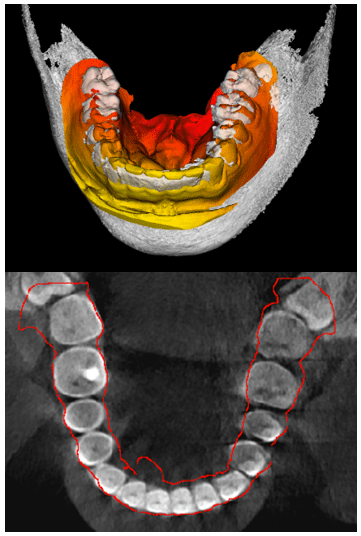}
                \includegraphics[width=1.3in]{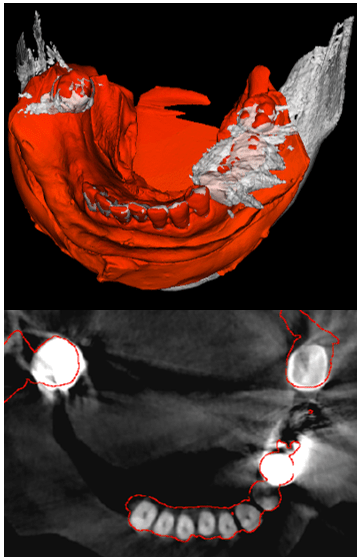}
                \includegraphics[width=1.3in]{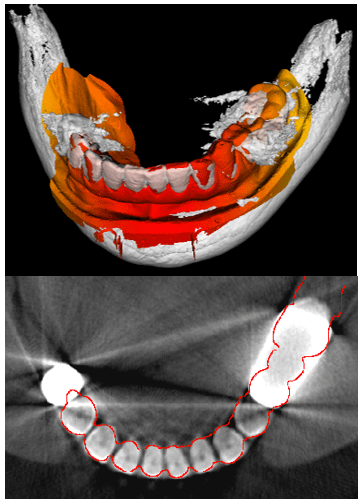}
            \captionof{subfigure}{Our method.}
        \end{minipage}
        \vfil
       
        \includegraphics[width=4in]{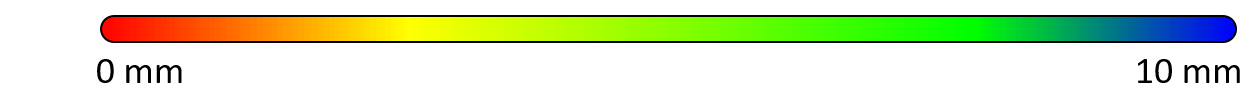}
        
    \caption{Visualization of the five methods. Each row represents different cases. (a) Three-points registration results. (b), (c), and (d) are the results of ICP \cite{besl1992method}, sparse ICP \cite{bouaziz2013sparse}, and Go-ICP \cite{yang2016go} methods, respectively, from the initial state (a). (e) Our proposed method with neither initial conditions nor manual points. The colors of the surfaces represent the distance between the corresponding result and the ground-truth. Explicit iso-surfaces are visualized together with the models for visual purposes. The three methods (b), (c), and (d) used the visualized iso-surfaces in their algorithms. Best viewed in color.}
    \label{fig:vis}
\end{figure*}

\subsection{Optimal Clusters}
The cluster-based fine registration algorithm is visualized in Fig. \ref{fig:cls}. As shown in Algorithm \ref{algo}, crown-prioritized clusters are generated for local optimization, and the three optimal clusters that have the most coherent local transformations are selected (Fig. \ref{fig:cls3}). The final registration procedure harnesses the three remaining optimal clusters as demonstrated in (\ref{eq:final}). The metal artifact regions whose crown boundaries are difficult to delineate are automatically neglected by the non-coherent transformation (Fig. \ref{fig:cls}). Consequently, the clusters that have relatively clear boundaries in the corresponding CT image were used for the final registration. We used $r=10$mm for all the experiments. The average number of initial clusters was 17, resulting in 68 clusters in total, including the stochastic clusters.
\par

\subsection{Performance Evaluation}
The evaluation results were compared with the ground-truth registration results provided by the experts in the field. In assessing the accuracy of the results, we used the distance between landmarks on surface models, which were also marked by the experts. Table \ref{table:results_stat3} shows the mean Euclidean distance errors of landmarks and the elapsed time for performing registration on 45 test subjects (i.e., 90 registration results in total). Distance errors were calculated based on the ten landmark points:
\begin{equation}
    d(\mathbf{l}_i; T_{\theta^*}) = \frac{1}{10}\sum_{i=0}^{10}{||\mathbf{l}_i-T_{\theta^*}(\mathbf{l}_i)||}_2,
\label{eq:rms}
\end{equation}
where $\mathbf{l}_i$ is a landmark point on the surface model and $T_{\theta^*}$ is the optimal transformation. We compared our proposed algorithm with three-points manual registration followed by ICP-based methods \cite{besl1992method, rusinkiewicz2001efficient, yang2016go, aiger20084, bouaziz2013sparse}. The methods are state-of-the-art approaches that are frequently used in industry and many types of research. To obtain the target points from a CT image, we first extracted iso-surface of the crown with the threshold value $1000$. It is well known that the bone structure can be segmented by $1000$ Hounsfield unit in CT images \cite{misch2007contemporary}. The three-points method is used as a base metric, and the basic ICP \cite{besl1992method}, sparse ICP \cite{bouaziz2013sparse}, and Go-ICP \cite{yang2016go} methods are used in the subsequent fine registration. We also experimented further by eliminating the stochastic clusters from our method.\par

The result indicates that our proposed method achieved the highest accuracy among all the state-of-the-art methods. The ICP-based methods showed poor accuracy compared to the three-points manual registration that was carefully performed by the experts. The degradation of the accuracy from ICP-based methods is reasonable because the iso-surfaces extracted from CT images were severely affected by noise (Fig. \ref{fig:vis}). Our method, on the other hand, successfully achieved the highest accuracy compared to other methods. The aforementioned experimental variant that skips the additional generation of clusters (i.e., proceeds without stochastic addition of clusters) also outperformed the ICP-based methods. Since the randomly added clusters have a higher chance of avoiding the metal artifact regions, the stochastic method improved the registration accuracy of our method. Stochastic rotations also aid the clusters to better match with the model in the initial state. The analysis of the local regions (i.e., local optimization and coherency analysis) makes our method relatively slower than the basic ICP method. However, unlike other methods, the complete automatization of the procedure and superior accuracy may compensate for the time complexity in dental applications.\par

Some of the samples are visualized in Fig. \ref{fig:vis}. Fig. \ref{fig:vis}a shows manual registration with three-points. The first row in Fig. \ref{fig:vis} clearly demonstrates that all the methods accurately perform matching when there is no metal artifact in the CT image. In the case of the crown regions with metal artifact, the ICP-based methods fail to precisely match the regions even when they are provided with the three-points as a prior (the second and third rows in Fig. \ref{fig:vis}b$\sim$d). The critical drawback of the ICP-based methods is that noisy region (i.e., metal artifact) strongly affects the target (i.e., iso-surface from CT image), and there is no explicit method to remove the metal noise in the iso-surface. Our proposed method (Fig. \ref{fig:vis}e), on the other hand, robustly matched the scanned models to CT images in every condition. The main factor of superior accuracy is the selective use of regions unaffected by noise and exclusion of noisy regions from the scope of consideration in the subsequent process (Fig. \ref{fig:cls}).\par

\subsection{Parameter Study}
\begin{figure}[t!]
    \centering
    \includegraphics[width=\linewidth]{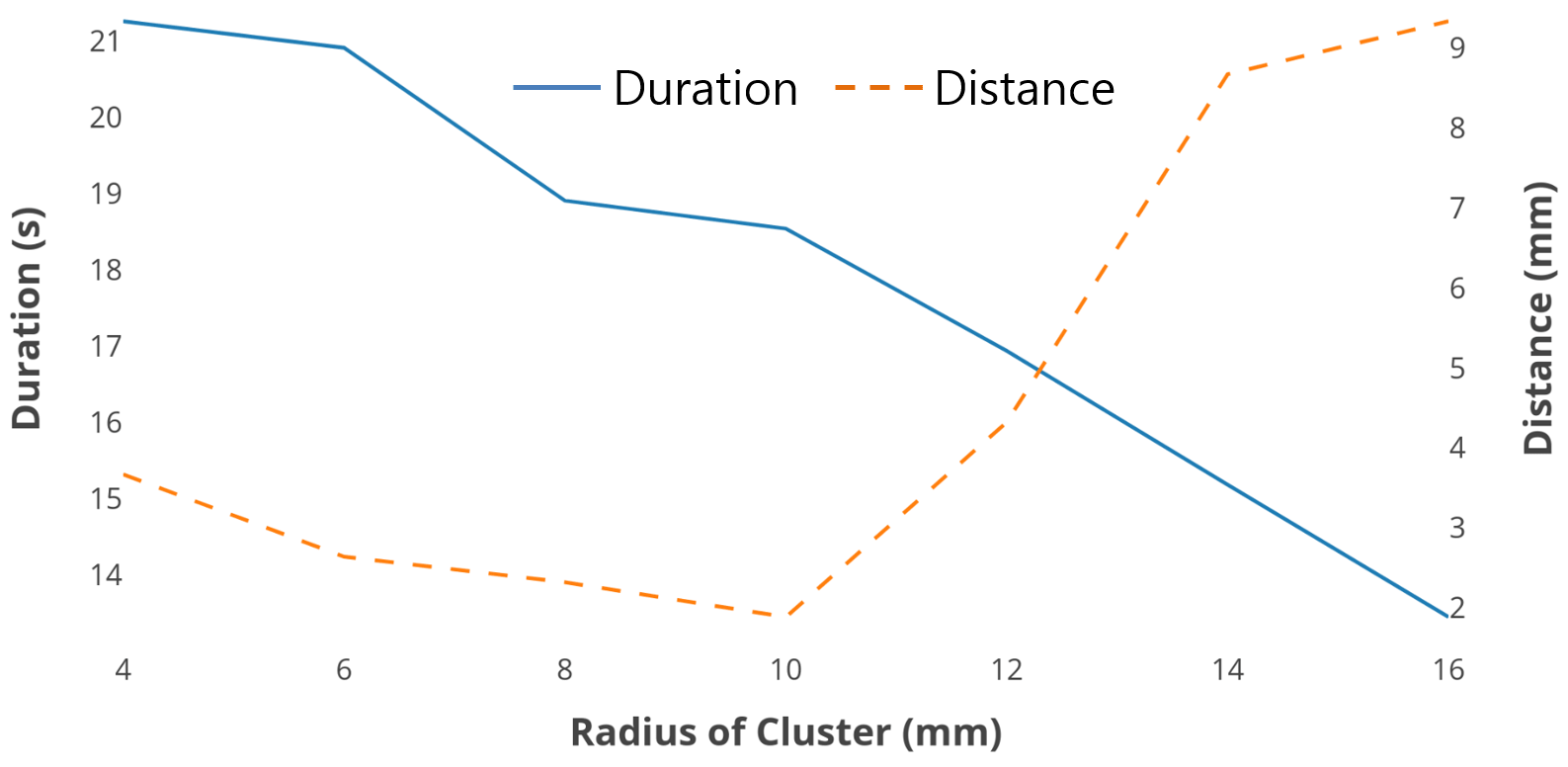}%
    \caption{The distance and run-time evaluation with respect to the radius of clusters. The mean Euclidean distance errors of landmarks are used for distance calculations.}
    \label{fig:plot}
\end{figure}

In this section, we study the radius parameter $r$ of a cluster. The size of the cluster affects the accuracy, run-time, and the number of clusters. Fig. \ref{fig:plot} shows the relationship between the radius, accuracy, and run-time. The optimum radius, which attained the highest accuracy, was 10mm (Fig. \ref{fig:plot}). The accuracy decreased steeply with bigger radius. It demonstrates that bigger clusters have a higher probability of containing noisy regions. The noise may either be metal artifacts or irrelevant portions that do not contain any corresponding features to the ones that exist in a CT scan. The duration time monotonically decreases as the radius increases from the optimum while losing the accuracy. It is obvious that a bigger radius leads to a decrease in the number of clusters and vice versa. The most time-consuming procedure of our proposed algorithm is the optimal cluster-based matching (i.e., local optimization). Local optimization with many small clusters is computationally costly (Fig. \ref{fig:plot}). In addition, iterative coherency analysis of transformation (\ref{eq:coherency}) is another cause of performance bottlenecks.\par

The cluster-based registration can be viewed as a special case of the ICP method. We iteratively removed outlying points (i.e., cluster) and performed correspondence matching via optimal vector alignments. We implicitly applied noise removal procedures by removing the clusters. The radius parameter does affect the performance of the proposed method. The results showed stable accuracy when using 10mm radius. The stability of the algorithm is mainly achieved by taking many cluster candidates into account (i.e., coherency evaluation in (\ref{eq:coherency})).

\section{Discussion and Conclusion}
Dental cone-beam CT image analysis is a challenging problem because of the large variety of cases and the presence of metal artifacts. To overcome the taxing manual labor of the registration between CT images and optically scanned models, we propose a fully automated CNN-based deep-pose estimations followed by cluster-based matching. Our work suggests the applicability of a neural network in simplifying the manual tasks involved in registration. Applying pose regressions with neural networks allowed the use of traditional methods in the following steps. As a result, we could avoid the pitfall of handling the entire registration task with complex neural networks. In fact, it is not feasible to solve the 3D registration problem directly with a neural network. The main limitations arise from the difficulties of generating the ground-truth data and unclear similarity criteria between images of different modality. As for fine registration, optimal cluster-based similarity matching was performed to accomplish confident region matching. Our method reduced the distance error by 30.77\% to 70\% when compared to other state-of-the-art registration methods. The primary factor for such improvement was the optimal local clusters that helped rule out the regions affected by metal artifacts.\par

The proposed algorithm is applicable to full-arch scanned models in a clinical practice. The performance of the proposed initial registration by deep regression networks shows results that are stable enough to be used for further fine registration process. Although working with full-arch scanned model is a common practice, the occasional use of partially scanned models may be problematic for the algorithm. This is because we employed the PCA method to obtain the axis for projection in the scanned model.\par




\ifCLASSOPTIONcaptionsoff
  \newpage
\fi

\bibliographystyle{IEEEtran}
\bibliography{MyBiB}

\end{document}